\documentclass[a4paper]{article}

\usepackage{INTERSPEECH2021}
\usepackage[utf8]{inputenc}
\usepackage{microtype}
\usepackage{graphicx}
\usepackage{subfigure}
\usepackage{booktabs}
\usepackage{varwidth}
\usepackage{url}
\usepackage{hyperref}

\usepackage{mathtools}
\usepackage{amsthm}
\usepackage{amsmath}
\usepackage[mathscr]{euscript}
\usepackage{bbm}
\usepackage{multirow}
\usepackage{wrapfig}
\let\vec\undefined
\usepackage{MnSymbol}
\usepackage{xpatch}
\usepackage{algorithm}
\usepackage{algpseudocode}

\let\Pr\relax
\DeclareMathOperator*{\Pr}{\mathbb{P}}

\DeclareMathOperator*{\E}{\mathbb E}

\DeclareMathOperator*{\argmin}{argmin}

\hypersetup{
  colorlinks   = true, 
  urlcolor     = blue, 
  linkcolor    = blue, 
  citecolor   = blue 
}

\newcommand{\cD}{\mathcal{D}}

\newcommand{\cO}{\mathcal{O}}

\newcommand{\cW}{\mathcal{W}}

\newcommand{\sD}{{\mathscr D}}

\newcommand{\sH}{{\mathscr H}}

\newcommand{\sL}{{\mathscr L}}

\newcommand{\sW}{{\mathscr W}}
\newcommand{\sX}{{\mathscr X}}
\newcommand{\sY}{{\mathscr Y}}

\newcommand{\bB}{{\mathbf B}}

\newcommand{\bL}{{\mathbf L}}

\newcommand{\bN}{{\mathbf N}}

\newcommand{\bR}{{\mathbf R}}

\renewcommand{\L}{\mathsf{L}}

\newcommand{\h}{\widehat}

\newcommand{\ignore}[1]{}

\newcommand\conf[1]{#1}

\newcommand{\afa}{\textsc{AgnosticFedAvg}}
\newcommand{\fedavg}{\textsc{FedAvg}}
\newcommand{\clientdomain}{\emph{client partition}}
\newcommand{\datadomain}{\emph{data partition}}
\newcommand{\Clientdomain}{\emph{Client partition}}
\newcommand{\Datadomain}{\emph{Data partition}}

\title{Communication-Efficient Agnostic Federated Averaging}
\name{Jae Ro$^1$, Mingqing Chen$^1$, Rajiv Mathews$^1$, Mehryar Mohri$^{1,2}$, Ananda Theertha Suresh$^1$}
\address{
  $^1$Google Inc.\\
  $^2$Courant Institute of Mathematical Sciences}
\email{\{jaero,mingqing,mathews,mohri,theertha\}@google.com}

\usepackage{graphicx, amsmath, amsfonts}

\begin{document}

\maketitle

\begin{abstract}
In distributed learning settings such as federated learning, the training algorithm can be potentially biased towards different clients. \cite{mohri2019agnostic} proposed a domain-agnostic learning algorithm, where the model is optimized for any target distribution formed by a mixture of the client distributions in order to overcome this bias. They further proposed an algorithm for the cross-silo federated learning setting, where the number of clients is small. We consider this problem in the cross-device setting, where the number of clients is much larger. We propose a communication-efficient distributed algorithm called \textsc{Agnostic Federated Averaging} (or \afa) to minimize the domain-agnostic objective proposed in \cite{mohri2019agnostic}, which is amenable to other private mechanisms such as secure aggregation. We highlight two types of naturally occurring domains in federated learning and argue that \afa\ performs well on both. To demonstrate the practical effectiveness of \afa, we report positive results for large-scale language modeling tasks in both simulation and live experiments, where the latter involves training language models for Spanish virtual keyboard for millions of user devices.
\end{abstract}

\section{Introduction}

In \emph{federated learning} (FL), a global model is trained on decentralized data from a large number of clients, which may be mobile phones, other edge devices, or sensors \cite{konevcny2016federated,konecny2016federated2, mcmahan2017communication}. The training data remains distributed over the clients, thus providing a layer of privacy during model training. However, FL also raises several types of issues, both practical and algorithmic, that have been the topic of multiple research efforts. This includes efficient communication strategies \cite{konevcny2016federated,konecny2016federated2, hamer2020fedboost, basu2020qsparse, haddadpour2021federated}, differential privacy algorithms \cite{AgarwalSureshYuKumarMcMahan2018, kairouz2021distributed}, lower bound guarantees for parallel stochastic optimization \cite{WoodworthWangSmithMcMahanSrebro2018}, better optimization algorithms \cite{li2018federated, karimireddy2020scaffold, NEURIPS2020_564127c0, karimireddy2020mime, acar2021federated, li2020unified}, and algorithms for adaptation, multi-task learning, and personalization \cite{smith2017federated, jiang2019improving, mansour2020three, kulkarni2020survey,  agarwal2020federated, mansour2021theory}. We refer readers to \cite{li2019federated} and \cite{kairouz2019advances} for a detailed literature survey on FL. FL is typically studied in two scenarios: \emph{cross-silo} and \emph{cross-device}. In cross-silo FL, the number of clients is small, where as in cross-device FL, the number of clients is very large and can be in the order of millions.  

Fairness is a key objective in general machine learning \cite{Bickel398,NIPS2016_9d268236} and especially FL \cite{abay2020mitigating,Li2020Fair}, where the network of clients can be massive and heterogeneous. Standard learning objectives in FL minimize the loss with respect to the uniform distribution over all samples. 
\cite{mohri2019agnostic} argued that, in many common instances, the uniform distribution is not the natural objective distribution as the data observed during training and inference in FL can differ. This is, in part, because models are typically trained on client devices under certain conditions (e.g. device is charging, is connected to an un-metered network, is idle, etc.), whereas during inference, these conditions need not be met.
 Hence it's risky to seek to minimize the expected loss with respect to a specific distribution. To overcome this, they proposed a new framework, \emph{agnostic federated learning}, where the centralized model is optimized for any possible target distribution formed by a mixture of the client distributions. Instead of optimizing for a specific distribution, which has the high risk of a mismatch with the target, they defined an agnostic and more risk-averse objective. They further showed generalization guarantees for this new objective and proposed a stochastic mirror descent type algorithm to minimize this objective.

However, their approach and algorithm did not address some key scenarios in FL. Firstly, their algorithm is feasible in the cross-silo setting, where the number of clients is small and the samples per client is large. However, in the cross-device setting, where the number of clients is very large, we argue that their model yields very loose generalization bounds. Secondly, their algorithm did not fully address the important communication bottleneck and decentralized data issues \cite{mcmahan2017communication} inherent in the cross-device FL setting. A straightforward implementation of their approach requires running a federated algorithm for a few hundred thousand rounds, which is not feasible in the cross-device setting. 

In this paper, we overcome these bottlenecks and 
propose a communication-efficient federated algorithm called \textsc{Agnostic Federated Averaging} (or \afa) to minimize the agnostic learning objective in the cross-device setting. \afa\ is not only communication-efficient, but also amenable to privacy preserving techniques such as secure aggregation \cite{bonawitz2017practical}.  The rest of the paper is organized as follows. In Section~\ref{sec:framework}, we state the notation and overview existing results, in Section~\ref{sec:new}, we define the framework, and 
in Section~\ref{sec:afa}, we propose our algorithm. Finally, in Section~\ref{sec:experiments}, we evaluate the proposed algorithm on different synthetic and live user datasets.

\begin{algorithm*}[t]
\caption{\afa}
\label{alg:AgnosticFedAvg}
\begin{varwidth}[t]{0.55\textwidth}
\begin{algorithmic}[1]
\Procedure{Server}{}
    \State $w_0 \in \sW$, $\lambda_0 \in \Delta_p$, $\bN_0 \in \mathbb{N}^{p}$
    \For{round $t = 1$ to $T$}
        \State $\alpha_t \gets \frac{\lambda_{t-1}}{\sum^{t-1}_{j=t-r} \bN_{j}/r}$
        \State $C_t \gets$ (random set of $c$ clients)
        \For{client $k \in C_t$}
            \State $w^k_t, \beta^k_t, \bL^k_t, \bN^k_t \gets \textsc{Client}(k, w_{t-1}, \alpha_t)$
        \EndFor
        \State $w_t \gets \sum_{k \in C_t}\beta^k_t w^k_t/\beta_t$
        \State $\bN_t \gets \sum_{k \in C_t} \bN^k_t$
        \State $\bL \gets \sum_{k \in C_t} \bL^k_t/\bN_t$
        \State $\lambda_t \gets \frac{\lambda_{t-1} \cdot \exp \left( \gamma_\lambda \bL \right) }{\sum^p_{i=1}     \lambda^i_{t-1} \cdot \exp \left( \gamma_\lambda \bL_i) \right)}$
    \EndFor
\EndProcedure
\end{algorithmic}
\end{varwidth}\quad              
\begin{varwidth}[t]{0.44\textwidth}
\begin{algorithmic}[1]
\Procedure{Client}{$k, w, \alpha$} \Comment{Run on client $k$}
    \For{domain $i = 1$ to $p$}
        \State $\bL^k_i \gets |S_k \cap \h{\sD}_i|\cdot  \L(w, S_k \cap \h{\sD}_i)$
        \State $\bN^k_i \gets |S_k \cap \h{\sD}_i|$
        \State $\beta^k \gets \sum^p_{i=1} \alpha_i  |S_k \cap \h{\sD}_i|$
    \EndFor
    \State $\bB \gets$ (split $S_k$ into batches of size $B$)
    \For{epoch $e = 1$ to $E$}
        \For{batch $b\in\bB$}
            \State 
         $ w \gets w - \gamma_w\nabla\left(\frac{\sum^p_{i=1} \alpha_i \sum_{j \in b \cap \h{\sD}_i} \L(w, x_j, y_j)}{\beta^k }\right)$
        \EndFor
    \EndFor
    \State \textbf{return} $w, \beta^k, \bL^k, \bN^k$
\EndProcedure
\end{algorithmic}
\end{varwidth}
\end{algorithm*}
\section{Preliminaries and Previous Work}
\label{sec:framework}
We start with some general notation and definitions. Let $\sX$ denote the input space and $\sY$ the output space. A distribution $\sD$ is a distribution over $\sX \times \sY$. 

We will primarily discuss a multi-class classification problem where $\sY$ is a finite set of classes, but much of our results can be extended straightforwardly to regression and other problems. The hypotheses we consider are of the form $h\colon \sX \to \Delta_\sY$, where $\Delta_\sY$ stands for the simplex over $\sY$. Thus, $h(x)$ is a probability distribution over the classes or categories that can be assigned to $x \in \sX$. We will denote by $\sH$ a family of such hypotheses $h$. We also denote by $\ell$ a loss function defined over $\Delta_\sY \times \sY$ taking non-negative values. The loss of $h \in \sH$ for a labeled sample $(x, y) \in \sX \times \sY$ is given by $\ell(h(x), y)$. One key example in applications is the cross-entropy loss, which is defined as
%
$\ell(h(x), y) = -\log (\Pr_{y' \sim h(x)}[y' = y]).$
%
We will denote by $\sL_\sD(h)$ the expected loss of a hypothesis $h$ with respect to a distribution $\sD$ over $\sX \times \sY$,
$
\sL_\sD(h) = \E_{(x, y) \sim \sD} [\ell(h(x), y)]
$
and by $h_\sD$ its minimizer: $h_\sD = \argmin_{h \in \sH} \sL_\sD(h)$. In standard learning scenarios,
the distribution $\sD$ is the test or target distribution, which typically coincides with the distribution of the training samples. However, in FL, this is often not the case.

In FL, the data is distributed across many heterogeneous clients and the data distribution is different for each client \cite{kairouz2019advances}. Let $q$ be the total number of clients. Let $\cD_k$ denote the data distribution for client $k$. The client does not have access to the true distribution $\cD_k$ and instead has access to $S_k$ where $S_k = ((x_{k, 1}, y_{k, 1}), \ldots, (x_{k, n_k}, y_{k, n_k})) \in (\sX \times \sY)^{n_k}$. Let $\h{\cD}_k$ denote the empirical distribution associated to sample $S_k$ of size $n_k$. A natural goal is to minimize the empirical risk on the average risk given by
\[
\sL_{\overline{U}}(h),
\]
where $\overline{U} = \frac{1}{q} \sum_k \h{\cD}_k$ is the uniform distribution over all clients data. However, as argued by \cite{mohri2019agnostic}, due to differences between the train and test distributions, minimizing this objective is risky. Hence, they proposed to minimize the loss on the worst case distribution. More concretely, for distributions $\cD_k$, $k = 1, \ldots, q$, let
$
\cD_\lambda =  \sum_{k = 1}^q \lambda_k \cD_k
$
for some $\lambda \in \Delta_q$, where $\Delta_q$ is the probability simplex over $q$ clients.
Thus, the learner minimizes the \emph{empirical agnostic loss} (or \emph{agnostic risk}) $\sL_{\cD_{\Delta_q}}(h)$ associated to a predictor $h \in \sH$ as
\begin{equation}
\label{eq:AgnosticRisk}
\sL_{\overline{\cD}_{\Delta_q}}(h) = \max_{\lambda \in {\Delta_q}} \sL_{\overline{ \cD}_\lambda}(h),
\end{equation}
where $\overline{ \cD}_\lambda = \sum_{k = 1}^q \lambda_k \h \cD_k$.  For simplicity, we allow any %
$\lambda \in \Delta_q$ in the above definition. 
However, the generalization bounds~\cite[Theorem 1]{mohri2019agnostic} depends on $\min_k n_k$, the minimum number of samples of any client. In the cross-device setting, this yields loose bounds as each client typically only has a few hundred samples.  Hence, instead of treating each client as a domain, we treat collections of clients or data pooled from clients as domains.

\section{Proposed Formulation}
\label{sec:new}
As stated before, treating each client as a separate domain yields loose generalization bounds. Hence, we treat collections of clients as domains, which naturally leads to two types of partitions. Let there be $p$ domains $\sD_1, \sD_2, \ldots, \sD_p$. 
\begin{enumerate}
    \item \Datadomain: Each client has data from one or more domains and domains represent different types of data. For example, for virtual keyboard applications~\cite{hard2018federated}, the domains could be the application source of client inputs, such as messaging, emails, or documents. In this case, the data distribution $\cD_k$ for client $k$ is given by
    \[
    \cD_k = \sum^p_{i=1} \lambda_i \sD_i,
    \]
    where $\sum^p_{i=1} \lambda_i = 1$ and $\lambda_i \geq 0$ for all $i \leq p$.
    \item \Clientdomain: Each client has data from exactly one domain and domains represent clusters of clients. For example, clustering clients based on their geographic location yields this domain type. In this case, the data distribution $\cD_k$ of client $k$ is given by
   $
    \cD_k = \sD_i \text { for some } i\leq p.
    $
\end{enumerate}

In both of the above formulations, even though there are $q$ different clients, the number of underlying distinct domains is $p$, which we argue is considerably smaller. Hence, we have a large number of samples from each of the domains and get strong generalization bounds.  Since each client distribution $\cD$ can be written as a linear combination of domain distributions,
\begin{equation}
    \label{eq:equivalent}
\max_{\lambda \in \Delta_q} \sL_{{\cD}_\lambda}(h) 
\leq 
\max_{\lambda \in \Delta_p} \sL_{{\sD}_\lambda}(h).
\end{equation}
However, we do not have access to the true domain distributions $\sD_i$ and instead have samples from $\h{\sD}_i$, where $\h{\sD}_i$ is the empirical distribution obtained by pooling all the data of domain $i$. Let $m_i$ be the number of samples in domain $i$.
By \eqref{eq:equivalent}, the true agnostic loss over clients is smaller than the
true agnostic loss over domains. Hence we propose to minimize the empirical agnostic loss over domains,
\[
\max_{\lambda \in \Delta_p} \sL_{\overline{\sD}_\lambda}(h),
\]
where $\overline{ \sD}_\lambda = \sum_{i = 1}^p \lambda_i \h \sD_i$.
The previous known generalization bounds from \cite[Lemma 3, Corollary 4]{mohri2019agnostic} yields the following generalization bound. Let $\epsilon > 0$. With probability at least $ 1- \delta$, for any client $k$ and any hypothesis $h$
\begin{align*}
\sL_{\cD_k}(h) & \leq \max_{\lambda \in \Delta_p} \sL_{\overline{\sD}_\lambda}(h) \\ &  + \sqrt{\frac{\ell_c}{\min_i m_i}} \left(  \sqrt{d \log \frac{\sum_i m_i}{d} + p \log \frac{1}{\epsilon \delta}}\right) + \epsilon,
\end{align*}
for some constant $\ell_c$ which depends on the maximum value of the loss and $d$ is the Vapnik–Chervonenkis (VC) dimension of the hypothesis class $\sH$.
The above generalization bound scales inversely with $\min_i m_i$, which is the minimum number of samples in any domain. Since the number of domains $p$ is small, as long as the domains are well-distributed, we would have a relatively large number of samples per domain and thus a favorable generalization bound in the cross-device setting. 

We now propose a communication-efficient algorithm to minimize agnostic loss~\eqref{eq:AgnosticRisk} in the cross-device setting.

\section{\afa}
\label{sec:afa}

\cite{mohri2019agnostic} showed that agnostic learning can be treated as a two-player game, where a learner tries to find the best hypothesis and an adversary tries to find the domain weights $\lambda$ that maximize the loss. They proposed a stochastic mirror descent algorithm and showed that the objective reaches the optimum value at a rate of $\cO(1/\sqrt{T})$ after $T$ rounds of training.
Similar to how \textsc{FederatedAveraging} (\fedavg) of \cite{mcmahan2017communication} is based on stochastic gradient descent (SGD) but is more communication-efficient, we propose \afa, that is based on  \cite{mohri2019agnostic} and is communication-efficient.
In fact, a direct implementation of \cite{mohri2019agnostic} would be infeasible in the cross-device setting, as the number of steps can be in the order of millions.
Furthermore, a direct implementation of \cite{mohri2019agnostic} requires the clients to reveal their domain to the server, which can be privacy-invasive. 
In contrast, the proposed 
algorithm 
\afa\
can be used with privacy preserving techniques such as secure aggregation. We design \afa\ with the following properties.
\begin{itemize}
    \item Each round of FL uses only a single round of transmission, which includes model download from the server to the clients and upload from the clients to the server.
    \item The clients train with multiple local SGD steps similar to \fedavg.
    \item The server does not have access to individual clients data but only aggregated statistics, making it compatible with other cryptographic techniques such as secure aggregation \cite{bonawitz2017practical}. This provides another layer of security and prevents the server from retrieving or rebuilding privacy-sensitive information from individual client parameter updates without additional side information.
\end{itemize}

Let $\sW$ be the set of parameters of the hypothesis class. The algorithm first initializes the weights to $w_0 \in \sW$, domain weights to $\lambda_0 \in \Delta_p$, and the number of examples per domain to $\bN_0 \in \mathbb{N}^p$, where $\bN_{t,i}$ denotes the number of samples for domain $i$ at round $t$ and $\bN^k$ denotes the number of samples for client $k$, split by domain.
We keep a sliding window of the number of examples per domain $\bN_t$ over the last $r$ training rounds. The algorithm uses learning rate $\gamma_\lambda$ for learning domain weights. 
In the following, let $\L$ denote the loss function as a function of hypothesis parameter $w$.

At each round of training $t$, the algorithm computes a scaling vector $\alpha_t$ by taking the ratio of domain weights $\lambda_{t-1}$ and the average number of samples per domain for the last $r$ rounds $\sum^{t-1}_{j=t-r} \bN_{j}/r$. The algorithm then selects $c$ clients randomly $C_t$ and sends the parameters $w_{t-1}$ and scaling vector $\alpha_t$ to each of them. 
First, each selected client $k$ computes the number of samples per domain $\bN^k$, initial loss per domain $\bL^k \in \bR^p$, and scaled client weight $\beta^k$ for their local dataset. Then, each client updates the parameters $w_{t-1}$ based on $\alpha_t$ and $\beta^k$ by running $E$ epochs of SGD with batch size $B$ and learning rate $\gamma_w$. Finally, the client transmits the updated parameters $w^k$, weight per client $\beta^k$, initial loss per domain $\bL^k$, and number of samples $\bN^k$ per domain back to the server. Since this is done using secure aggregation, the server only observes the total number of samples $\bN$ and loss $\bL$ per domain across clients. The server then computes the new parameters $w_t$ by averaging the client updates weighted by $\beta^k_t$ and does an exponentiated gradient (EG) step for the domain weights $\lambda_t$,
\begin{align*}
\lambda_t 
&= \frac{\lambda_{t-1} \cdot \exp \left( \gamma_\lambda \bL \right) }{\sum^p_{i=1} \lambda^i_{t-1} \cdot \exp \left( \gamma_\lambda \bL_i) \right)}.
\end{align*}
If a round does not have any samples from a particular domain, we set $\bL$ to zero for that round. This process is repeated for $T$ rounds. The complete pseudo-code is given in Algorithm~\ref{alg:AgnosticFedAvg}.

To see why the above algorithm aims to minimize the agnostic loss, consider the weighted average of all the client losses
\begin{align*}
&\sum_{k \in C_t}\beta^k_{t} \frac{\sum^p_{i=1} \alpha^i_t \sum_{j \in S_k \cap \h{\sD}_i} \L(w, x_j, y_j)}{\sum^p_{i=1} \alpha^i_t  |S_k \cap \h{\sD}_i| } \\
&= \sum_{k \in C_t}  \sum^p_{i=1} \alpha^i_t \sum_{j \in S_k \cap \h{\sD}_i}  \L(w, x_j, y_j) \\
&= \sum^p_{i=1} \alpha^i_t \sum_{k \in C_t}  \sum_{j \in S_k \cap \h{\sD}_i}  \L(w, x_j, y_j) \\
&\stackrel{(a)}{\approx} \sum^p_{i=1} \lambda^i_t \frac{\sum_{k \in C_t}  \sum_{j \in S_k \cap \h{\sD}_i}  \L(w, x_j, y_j)}{N_i} 
= \sum^p_{i=1} \lambda^i_t \L_i(w),
\end{align*}
where $\L_i(w)$ is the average loss for domain $i$ and $N_i$ is the number of samples in domain $i$ from the selected clients at round $t$. The approximation $(a)$ assumes that the moving average of $\bN_{t, i}$ is close to the number of samples in domain $i$ from the selected clients at round $t$. Thus, \afa\ aims to minimize the domain agnostic objective defined in~\eqref{eq:AgnosticRisk}. We further note that by using secure aggregation \cite{bonawitz2017practical}, the server only observes aggregated statistics rather than learning domains or gradients of individual clients and provides an additional layer of privacy.


The communication costs of \fedavg\ and \afa\ are given in Table~\ref{table:comm-efficiency}.
For a given round $t$, \afa\ adds a small additional cost on top of the communication cost of \fedavg, as the number of domains $p$ is typically much smaller than the number of model parameters $|\cW|$.  Furthermore, in practice, \afa\ can use fewer communication rounds than \fedavg, thereby reducing or eliminating this overhead entirely (Appendix~\ref{app:comm-efficiency}).

\begin{table}[t]
\caption{Total communication cost in number of parameters per round, where $|\cW|$ is the number of parameters in the model.}
\centering
\small{
\begin{tabular}{ll}
algorithm           &  number of parameters per round \\ \hline
\fedavg             & $2 c \cdot |\cW|$  \\
\afa                & $2 c \cdot |\cW| + 4 c \cdot p$            
\end{tabular}
\label{table:comm-efficiency}
}
\vskip -.2in
\end{table}

\section{Experiments}
\label{sec:experiments}

\begin{table*}[t]
\caption{Perplexity and in-vocab-accuracy for Stack Overflow test dataset with the standard deviation for three trials in parentheses. \afa\ attains lowest perplexity and highest in-vocab-accuracy for the harder domain \textbf{answer}.}
\vskip -.1in
\centering
\small{
\begin{tabular}{ccccccc}
algorithm           & \multicolumn{2}{c}{\textbf{answer}}                   & \multicolumn{2}{c}{question}                          & \multicolumn{2}{c}{difference}    \\ \hline
                    & perp.                     & acc.                      & perp.                     & acc.                      & perp.             & acc.      \\
\fedavg\ (uniform)  & $53.1 \,(.06)$            & $24.5\, (.02)$            & $\mathbf{43.4 \,(.23)}$   & $\mathbf{27.5 \,(.06)}$   & $9.7$             & $3.0$     \\
\fedavg\ (answer)   & $52.9\, (.15)$            & $24.9\, (.02)$            & $64.2\, (.39)$            & $21.2\, (.14)$            & $11.3$            & $3.7$     \\
\afa                & $\mathbf{51.9 \,(.22)}$   & $\mathbf{25.1 \,(.004)}$  & $52.7 \,(.36)$            & $23.5\, (.05)$            & $\mathbf{0.8}$    & $\mathbf{1.6}$    
\end{tabular}
\label{table:stackoverflow-results}
}
\vskip -.05in
\end{table*}

\begin{table}[t]
\caption{Perplexity and in-vocab-accuracy for Spanish virtual keyboard. \afa\ attains lowest perplexity  for the harder domain \textbf{es-AR}.}
\centering
\vskip -.1in
\small{
\begin{tabular}{@{\hspace{0cm}}l@{\hspace{.2cm}}cc@{\hspace{.2cm}}cc@{\hspace{.2cm}}cc@{\hspace{0cm}}}
algorithm           & \multicolumn{2}{c}{\textbf{es-AR}}    & \multicolumn{2}{c}{es-419$^*$}            & \multicolumn{2}{c}{es-US} \\ \hline
                    & perp.             & acc.              & perp.             & acc.              & perp.             & acc.  \\
\fedavg\ (uniform)  & $56.0$            & $11.9$            & $\mathbf{50.5}$   & $\mathbf{11.3}$   & $\mathbf{44.2}$   & $\mathbf{10.5}$   \\
\fedavg\ (es-AR)    & $55.0$            & $12.1$            & $62.2$            & $10.2$            & $52.8$            & $9.5$ \\
\afa                & $\mathbf{53.4}$   & $\mathbf{12.3}$   & $60.6$            & $10.2$            & $52.2$            & $9.6$            
\end{tabular}
\label{table:keyboard-results}
}
\vskip -.05in
\end{table}

\begin{table}[t]
\caption{Statistics per domain in the Stack Overflow dataset.}
\vskip -.1in
\centering
\small{
\begin{tabular}{cccc}
                & train & held-out & test     \\ \hline
clients           & 342K  & 38.8K    & 204K      \\
sentences       & 136M  & 16.5M    & 16.6M     \\
answers         & 78.0M & 9.33M    & 9.07M    \\    
questions       & 57.8M & 7.17M    & 7.52M
\end{tabular}
\label{table:stackoverflow-data}
}
\vskip -.05in
\end{table}
We report the results for the English Stack Overflow language model simulation task and a Spanish language modeling live experiment for millions of virtual keyboard user devices. 
We implemented all algorithms and experiments using the open-source FedJAX \cite{fedjax2020github} and TensorFlow Federated \cite{tff2019} libraries.  For all experiments, we compare three algorithms:
\begin{itemize}
 {\setlength\itemindent{-15pt} \item  \textbf{\fedavg\ (uniform)}: Trained uniformly on all available data.}
 {\setlength\itemindent{-15pt} \item \textbf{\fedavg\ (target-only)}: Trained only on data from the target.}
 {\setlength\itemindent{-15pt} \item \textbf{\afa}: Trained on all available data.}
\end{itemize}
We demonstrate that \afa\ attains a lower perplexity  compared to \fedavg\ (uniform) and \fedavg\ (target-only) for both the experiments on the harder domain: answer domain for Stack Overflow and es-AR for the Spanish language model.

To verify that \afa\ correctly minimizes the domain agnostic objective and to showcase its effectiveness on non-language tasks, we also include experiments on a synthetic toy regression example and the EMNIST-62 image recognition task in Appendices~\ref{app:toy-regression} and~\ref{app:emnist}, respectively.

\subsection{Stack Overflow Language Model}
\label{stackoverflowsection}

We consider the language model task for the Stack Overflow dataset from \cite{reddi2020adaptive}. This dataset contains two domains, questions and answers, from the Stack Overflow forum grouped by client ids. This corresponds to the \datadomain\ domain type since an individual client can post both questions and answers. Table~\ref{table:stackoverflow-data} summarizes the statistics per domain.

We match the model and training setup from \cite{reddi2020adaptive} and train a single layer LSTM language model over the top $10$K words with an Adam server optimizer and $50$ clients participating per training round for $1500$ rounds. For \afa, we use the same set up with domain weight learning rate $0.005$. 

For the Stack Overflow experiments, we report perplexity and in-vocab-accuracy, where in-vocab-accuracy is the number of correct predictions, without UNK (out-of-vocabulary) or EOS (end-of-sentence) tokens, divided by the number of words without the EOS token. Defining in-vocab-accuracy this way allows valid comparisons for different vocabulary sizes. The results are in Table~\ref{table:stackoverflow-results}. For the baseline \fedavg\ (uniform), of the two domains, the answer domain is harder and has higher perplexity and lower accuracy. Given this, we also train an additional baseline \fedavg\ (answer) on answer examples only. While \fedavg\ (answer) does improve answer performance over \fedavg\ (uniform), it results in significantly worse question performance. However, \afa\ outperforms both \fedavg\ (uniform) and \fedavg\ (answer) on the answers domain, while also significantly decreasing the performance disparity between answers and questions. This suggests that there could be important features in the questions that can augment performance on answers that are leveraged by \afa\ but aren't optimally weighted in \fedavg\ (uniform) or are completely ignored in \fedavg\ (answer).

\subsection{Spanish Virtual Keyboard Language Model}

We further use \afa\ to train a Coupled Input and Forget Gate (CIFG)~\cite{greff2017lstm} language model for Spanish on virtual keyboard client devices.
We follow the same settings and FL requirements for client participation as \cite{hard2018federated}.
We consider three domains based on the Spanish locales: es-US for US, es-AR for Argentina, and a subset of countries belonging to es-419\footnote{Defined by UN M.49 region code. We use ``es-419$^*$'' to denote these countries.}.
Since each user device falls in a single region, this task corresponds to the \clientdomain. 

Similar to Section~\ref{stackoverflowsection}, we report perplexity and in-vocab-accuracy. For all algorithms, we use the momentum server optimizer, using Nesterov accelerated gradient~\cite{nesterov}, and $500$ clients participating per training round for $3000$ rounds.
Over the course of training, approximately 141 million sentences are processed by 1.5 million clients. 
The results are in Table~\ref{table:keyboard-results}. For the baseline \fedavg\ (uniform), of the three languages, es-AR has the worst perplexity. Similar to Section~\ref{stackoverflowsection}, training \fedavg\ (es-AR) on es-AR clients only improves es-AR performance over \fedavg\ (uniform) but also results in much worse performance for es-US and es-419$^*$. Again, \afa\ improves the perplexity and accuracy on es-AR over \fedavg\ (uniform) while also decreasing the regression on es-US and es-419$^*$ when compared to \fedavg\ (es-AR).

\section{Conclusion}

We presented an algorithmic study of domain agnostic learning in the cross-device FL setting. We also examined the two types of naturally occurring domains in FL: \datadomain\ and \clientdomain\ and provided example learning tasks for both in large-scale language modeling. Finally, we defined \afa, a communication-efficient federated algorithm that aims to minimize the domain agnostic objective proposed in \cite{mohri2019agnostic} and can provide additional security using secure aggregation and demonstrated its practical effectiveness in simulations and real live experiments. We hope that our efforts will spur further studies into improving the practical efficiency of FL algorithms.

\section{Acknowledgements}

We thank our colleagues Kaan Ege Ozgun, Gary Sivek, Zachary Garrett, and Keith Rush for their help with training and experiment infrastructure, and Andrew Hard, Sean Campbell, and Françoise Beaufays for their helpful comments and discussions.

\newpage

\bibliographystyle{IEEEtran}
\bibliography{references}

\begin{thebibliography}{10}
\providecommand{\url}[1]{#1}
\csname url@samestyle\endcsname
\providecommand{\newblock}{\relax}
\providecommand{\bibinfo}[2]{#2}
\providecommand{\BIBentrySTDinterwordspacing}{\spaceskip=0pt\relax}
\providecommand{\BIBentryALTinterwordstretchfactor}{4}
\providecommand{\BIBentryALTinterwordspacing}{\spaceskip=\fontdimen2\font plus
\BIBentryALTinterwordstretchfactor\fontdimen3\font minus
  \fontdimen4\font\relax}
\providecommand{\BIBforeignlanguage}[2]{{%
\expandafter\ifx\csname l@#1\endcsname\relax
\typeout{** WARNING: IEEEtran.bst: No hyphenation pattern has been}%
\typeout{** loaded for the language `#1'. Using the pattern for}%
\typeout{** the default language instead.}%
\else
\language=\csname l@#1\endcsname
\fi
#2}}
\providecommand{\BIBdecl}{\relax}
\BIBdecl

\bibitem{mohri2019agnostic}
M.~Mohri, G.~Sivek, and A.~T. Suresh, ``Agnostic federated learning,'' in
  \emph{International Conference on Machine Learning}, 2019, pp. 4615--4625.

\bibitem{konevcny2016federated}
J.~Kone{\v{c}}n{\`y}, H.~B. McMahan, F.~X. Yu, P.~Richt{\'a}rik, A.~T. Suresh,
  and D.~Bacon, ``Federated learning: Strategies for improving communication
  efficiency,'' \emph{arXiv preprint arXiv:1610.05492}, 2016.

\bibitem{konecny2016federated2}
J.~Kone{\v{c}}n{\`y}, H.~B. McMahan, D.~Ramage, and P.~Richt{\'a}rik,
  ``Federated optimization: Distributed machine learning for on-device
  intelligence,'' \emph{arXiv preprint arXiv:1610.02527}, 2016.

\bibitem{mcmahan2017communication}
B.~McMahan, E.~Moore, D.~Ramage, S.~Hampson, and B.~A. y~Arcas,
  ``Communication-efficient learning of deep networks from decentralized
  data,'' in \emph{Artificial Intelligence and Statistics}.\hskip 1em plus
  0.5em minus 0.4em\relax PMLR, 2017, pp. 1273--1282.

\bibitem{hamer2020fedboost}
J.~Hamer, M.~Mohri, and A.~T. Suresh, ``Fedboost: A communication-efficient
  algorithm for federated learning,'' in \emph{International Conference on
  Machine Learning}.\hskip 1em plus 0.5em minus 0.4em\relax PMLR, 2020, pp.
  3973--3983.

\bibitem{basu2020qsparse}
D.~Basu, D.~Data, C.~Karakus, and S.~N. Diggavi, ``Qsparse-local-sgd:
  Distributed sgd with quantization, sparsification, and local computations,''
  \emph{IEEE Journal on Selected Areas in Information Theory}, vol.~1, no.~1,
  pp. 217--226, 2020.

\bibitem{haddadpour2021federated}
F.~Haddadpour, M.~M. Kamani, A.~Mokhtari, and M.~Mahdavi, ``Federated learning
  with compression: Unified analysis and sharp guarantees,'' in
  \emph{International Conference on Artificial Intelligence and
  Statistics}.\hskip 1em plus 0.5em minus 0.4em\relax PMLR, 2021, pp.
  2350--2358.

\bibitem{AgarwalSureshYuKumarMcMahan2018}
N.~Agarwal, A.~T. Suresh, F.~X. Yu, S.~Kumar, and B.~McMahan, ``{cpSGD}:
  Communication-efficient and differentially-private distributed {SGD},'' in
  \emph{Proceedings of NeurIPS}, 2018, pp. 7575--7586.

\bibitem{kairouz2021distributed}
P.~Kairouz, Z.~Liu, and T.~Steinke, ``The distributed discrete gaussian
  mechanism for federated learning with secure aggregation,'' \emph{arXiv
  preprint arXiv:2102.06387}, 2021.

\bibitem{WoodworthWangSmithMcMahanSrebro2018}
B.~E. Woodworth, J.~Wang, A.~D. Smith, B.~McMahan, and N.~Srebro, ``Graph
  oracle models, lower bounds, and gaps for parallel stochastic optimization,''
  in \emph{Proceedings of NeurIPS}, 2018, pp. 8505--8515.

\bibitem{li2018federated}
T.~Li, A.~K. Sahu, M.~Zaheer, M.~Sanjabi, A.~Talwalkar, and V.~Smith,
  ``Federated optimization in heterogeneous networks,'' \emph{arXiv preprint
  arXiv:1812.06127}, 2018.

\bibitem{karimireddy2020scaffold}
S.~P. Karimireddy, S.~Kale, M.~Mohri, S.~Reddi, S.~Stich, and A.~T. Suresh,
  ``Scaffold: Stochastic controlled averaging for federated learning,'' in
  \emph{International Conference on Machine Learning}.\hskip 1em plus 0.5em
  minus 0.4em\relax PMLR, 2020, pp. 5132--5143.

\bibitem{NEURIPS2020_564127c0}
\BIBentryALTinterwordspacing
J.~Wang, Q.~Liu, H.~Liang, G.~Joshi, and H.~V. Poor, ``Tackling the objective
  inconsistency problem in heterogeneous federated optimization,'' in
  \emph{Advances in Neural Information Processing Systems}, H.~Larochelle,
  M.~Ranzato, R.~Hadsell, M.~F. Balcan, and H.~Lin, Eds., vol.~33.\hskip 1em
  plus 0.5em minus 0.4em\relax Curran Associates, Inc., 2020, pp. 7611--7623.
  [Online]. Available:
  \url{https://proceedings.neurips.cc/paper/2020/file/564127c03caab942e503ee6f810f54fd-Paper.pdf}
\BIBentrySTDinterwordspacing

\bibitem{karimireddy2020mime}
S.~P. Karimireddy, M.~Jaggi, S.~Kale, M.~Mohri, S.~J. Reddi, S.~U. Stich, and
  A.~T. Suresh, ``Mime: Mimicking centralized stochastic algorithms in
  federated learning,'' \emph{arXiv preprint arXiv:2008.03606}, 2020.

\bibitem{acar2021federated}
D.~A.~E. Acar, Y.~Zhao, R.~M. Navarro, M.~Mattina, P.~N. Whatmough, and
  V.~Saligrama, ``Federated learning based on dynamic regularization,'' in
  \emph{International Conference on Learning Representations}, 2021.

\bibitem{li2020unified}
Z.~Li and P.~Richt{\'a}rik, ``A unified analysis of stochastic gradient methods
  for nonconvex federated optimization,'' \emph{arXiv preprint
  arXiv:2006.07013}, 2020.

\bibitem{smith2017federated}
V.~Smith, C.-K. Chiang, M.~Sanjabi, and A.~Talwalkar, ``Federated multi-task
  learning,'' in \emph{Proceedings of the 31st International Conference on
  Neural Information Processing Systems}, 2017, pp. 4427--4437.

\bibitem{jiang2019improving}
Y.~Jiang, J.~Kone{\v{c}}n{\`y}, K.~Rush, and S.~Kannan, ``Improving federated
  learning personalization via model agnostic meta learning,'' \emph{arXiv
  preprint arXiv:1909.12488}, 2019.

\bibitem{mansour2020three}
Y.~Mansour, M.~Mohri, J.~Ro, and A.~T. Suresh, ``Three approaches for
  personalization with applications to federated learning,'' \emph{arXiv
  preprint arXiv:2002.10619}, 2020.

\bibitem{kulkarni2020survey}
V.~Kulkarni, M.~Kulkarni, and A.~Pant, ``Survey of personalization techniques
  for federated learning,'' in \emph{2020 Fourth World Conference on Smart
  Trends in Systems, Security and Sustainability (WorldS4)}.\hskip 1em plus
  0.5em minus 0.4em\relax IEEE, 2020, pp. 794--797.

\bibitem{agarwal2020federated}
A.~Agarwal, J.~Langford, and C.-Y. Wei, ``Federated residual learning,''
  \emph{arXiv preprint arXiv:2003.12880}, 2020.

\bibitem{mansour2021theory}
Y.~Mansour, M.~Mohri, J.~Ro, A.~T. Suresh, and K.~Wu, ``A theory of
  multiple-source adaptation with limited target labeled data,'' in
  \emph{International Conference on Artificial Intelligence and
  Statistics}.\hskip 1em plus 0.5em minus 0.4em\relax PMLR, 2021, pp.
  2332--2340.

\bibitem{li2019federated}
T.~Li, A.~K. Sahu, A.~Talwalkar, and V.~Smith, ``Federated learning:
  Challenges, methods, and future directions,'' \emph{IEEE Signal Processing
  Magazine}, vol.~37, no.~3, pp. 50--60, 2020.

\bibitem{kairouz2019advances}
H.~B. McMahan \emph{et~al.}, ``Advances and open problems in federated
  learning,'' \emph{Foundations and Trends{\textregistered} in Machine
  Learning}, vol.~14, no.~1, 2021.

\bibitem{Bickel398}
P.~J. Bickel, E.~A. Hammel, and J.~W. O{\textquoteright}Connell, ``Sex bias in
  graduate admissions: Data from {B}erkeley,'' \emph{Science}, vol. 187, no.
  4175, pp. 398--404, 1975.

\bibitem{NIPS2016_9d268236}
\BIBentryALTinterwordspacing
M.~Hardt, E.~Price, E.~Price, and N.~Srebro, ``Equality of opportunity in
  supervised learning,'' in \emph{Advances in Neural Information Processing
  Systems}, D.~Lee, M.~Sugiyama, U.~Luxburg, I.~Guyon, and R.~Garnett, Eds.,
  vol.~29.\hskip 1em plus 0.5em minus 0.4em\relax Curran Associates, Inc.,
  2016. [Online]. Available:
  \url{https://proceedings.neurips.cc/paper/2016/file/9d2682367c3935defcb1f9e247a97c0d-Paper.pdf}
\BIBentrySTDinterwordspacing

\bibitem{abay2020mitigating}
A.~Abay, Y.~Zhou, N.~Baracaldo, S.~Rajamoni, E.~Chuba, and H.~Ludwig,
  ``Mitigating bias in federated learning,'' 2020.

\bibitem{Li2020Fair}
\BIBentryALTinterwordspacing
T.~Li, M.~Sanjabi, A.~Beirami, and V.~Smith, ``Fair resource allocation in
  federated learning,'' in \emph{International Conference on Learning
  Representations}, 2020. [Online]. Available:
  \url{https://openreview.net/forum?id=ByexElSYDr}
\BIBentrySTDinterwordspacing

\bibitem{bonawitz2017practical}
K.~Bonawitz, V.~Ivanov, B.~Kreuter, A.~Marcedone, H.~B. McMahan, S.~Patel,
  D.~Ramage, A.~Segal, and K.~Seth, ``Practical secure aggregation for
  privacy-preserving machine learning,'' in \emph{Proceedings of the 2017 ACM
  SIGSAC Conference on Computer and Communications Security}.\hskip 1em plus
  0.5em minus 0.4em\relax ACM, 2017, pp. 1175--1191.

\bibitem{hard2018federated}
A.~Hard, K.~Rao, R.~Mathews, F.~Beaufays, S.~Augenstein, H.~Eichner, C.~Kiddon,
  and D.~Ramage, ``Federated learning for mobile keyboard prediction,''
  \emph{arXiv preprint arXiv:1811.03604}, 2018.

\bibitem{fedjax2020github}
\BIBentryALTinterwordspacing
J.~H. Ro, A.~T. Suresh, and K.~Wu, ``{F}ed{JAX}: Federated learning simulation
  with {JAX},'' 2020. [Online]. Available:
  \url{http://github.com/google/fedjax}
\BIBentrySTDinterwordspacing

\bibitem{tff2019}
\BIBentryALTinterwordspacing
K.~Bonawitz, H.~Eichner, W.~Grieskamp, D.~Huba, A.~Ingerman, V.~Ivanov,
  C.~Kiddon, J.~Konecn{\'{y}}, S.~Mazzocchi, H.~B. McMahan, T.~V. Overveldt,
  D.~Petrou, D.~Ramage, and J.~Roselander, ``Towards federated learning at
  scale: System design,'' \emph{CoRR}, vol. abs/1902.01046, 2019. [Online].
  Available: \url{http://arxiv.org/abs/1902.01046}
\BIBentrySTDinterwordspacing

\bibitem{reddi2020adaptive}
S.~Reddi, Z.~Charles, M.~Zaheer, Z.~Garrett, K.~Rush, J.~Konečný, S.~Kumar,
  and H.~B. McMahan, ``Adaptive federated optimization,'' 2020.

\bibitem{greff2017lstm}
K.~Greff, R.~K. Srivastava, J.~Koutn{\'\i}k, B.~R. Steunebrink, and
  J.~Schmidhuber, ``{LSTM}: A search space odyssey,'' \emph{IEEE transactions
  on neural networks and learning systems}, vol.~28, no.~10, pp. 2222--2232,
  2017.

\bibitem{nesterov}
\BIBentryALTinterwordspacing
Y.~Nesterov, ``A method for solving the convex programming problem with
  convergence rate $o(1/{k}^{2})$,'' \emph{Dokl. Akad. Nauk SSSR}, vol. 269,
  pp. 543--547, 1983. [Online]. Available:
  \url{https://ci.nii.ac.jp/naid/10029946121/en/}
\BIBentrySTDinterwordspacing

\bibitem{caldas2018leaf}
S.~Caldas, S.~M.~K. Duddu, P.~Wu, T.~Li, J.~Kone{\v{c}}n{\`y}, H.~B. McMahan,
  V.~Smith, and A.~Talwalkar, ``Leaf: A benchmark for federated settings,''
  \emph{arXiv preprint arXiv:1812.01097}, 2018.

\end{thebibliography}

\newpage
\appendix

\conf{
\onecolumn
\begin{center}
    {\Large{ Supplementary material: Communication-Efficient Agnostic Federated Averaging}}
\end{center}
}

\section{Communication-Efficiency}
\label{app:comm-efficiency}

\begin{figure}[h]
    \centering
    \caption{Left: In-vocab-accuracy of answers over training rounds for the Stack Overflow experiments. Right: In-vocab-accuracy of es-AR over training rounds for the Spanish language modeling experiments.}
    \begin{tabular}{c}
        \includegraphics[width=0.7\columnwidth]{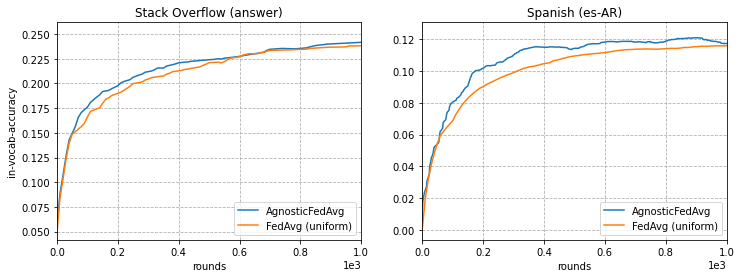}
    \end{tabular}
    
    \label{fig:lm-comm-results}
\end{figure}

We examine the communication efficiency of \afa\ by comparing the model performance between \fedavg\ and \afa\ throughout training. Figure~\ref{fig:lm-comm-results} reports in-vocab-accuracy for the harder target domain over communication rounds for the Stack Overflow and Spanish virtual keyboard language models. Within the first $1000$ rounds, \afa\ achieves a higher in-vocab-accuracy earlier compared to \fedavg\ in the harder domains, answers for Stack Overflow and es-AR for Spanish virtual keyboard.  Thus, although there is a small additional overhead introduced by \afa, the actual communication cost can be much lower than \fedavg, since \afa\ converges in significantly fewer rounds for the harder domain.

\section{Toy Regression}
\label{app:toy-regression}

\begin{figure}[h]
    \centering
    \caption{Experiments on the toy regression dataset. Left: Learned value over training rounds. Right: Domain weights over training rounds.}
    \begin{tabular}{cc}
        \includegraphics[width=0.3\columnwidth]{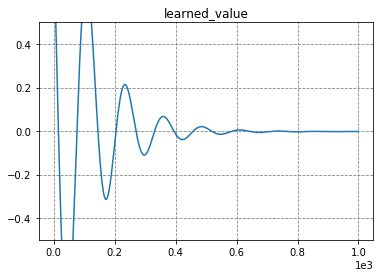} &                      \includegraphics[width=0.3\columnwidth]{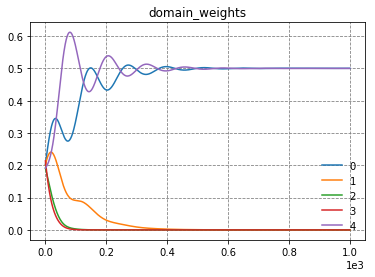} \\   
    \end{tabular}
    \label{fig:toy-results}
\end{figure}

We first evaluate \afa\ on a toy regression task to ensure its correctness. We consider a simple regression example, where each domain is a set of random points in $\bR$. Let each domain $i$, be a set of points $x_{i,1}, x_{i,2},\ldots, x_{i,m}$ in $\bR$. Further, let $c_i = \frac{1}{m} \sum^m_{j=1}x_{i,j}$ be the center of these points. We distribute these points on $50$ clients randomly. The goal is to find the point that minimizes the maximum distance to all the domain centers i.e.,
\[
\min_{w \in \bR} \max_{i \leq p} || c_i - w||^2.
\]
It is easy to see that 
\[
\min_{w \in \bR} \max_{i \leq p} || c_i - w||^2 = \min_{w} \max_{\lambda \in \Delta_p} \sum^p_{i=1} \lambda_i || c_i - w||^2,
\]
thus we maximize the latter objective by \afa. We choose points such that the true answer is $0$ and plot the performance of \afa\ for 5 domains in Figure~\ref{fig:toy-results}. As expected, \afa\ converges to the true solution within $1000$ rounds.

\section{EMNIST-62 Image Recognition}
\label{app:emnist}

We consider the image recognition task for the EMNIST-62 dataset \cite{caldas2018leaf} provided by TensorFlow Federated~\cite{tff2019}. This dataset consists of $3400$ writers and their writing samples which are one of $62$ classes (alphanumeric). According to the original NIST source documentation\footnote{https://s3.amazonaws.com/nist-srd/SD19/sd19\_users\_guide\_edition\_2.pdf}, the writers come from two distinct sources: high school and census field. This corresponds to the \clientdomain\ domain type since a given client can only belong to a single domain. Table~\ref{table:emnist-data} summarizes the statistics on the number of clients and examples per domain.

We match the model and training setup from \cite{reddi2020adaptive} and train a convolution neural net with an Adam server optimizer and $10$ clients participating per training round for $1500$ rounds. For \afa, we use the same set up with domain weight learning rate $0.01$. \cite{reddi2020adaptive} provides a comprehensive overview over different server optimizer varieties and their respective performances. For our experiments, we use the Adam server optimizer as it was shown to produce the highest accuracy.

The results are in Table~\ref{table:emnist-results}. For the baseline \fedavg\ (uniform), of the two domains, the high school domain is harder and has lower accuracy, most likely because it has fewer clients and training examples. In light of this, we also train \fedavg (high school) only on clients from the high school domain. While \fedavg\ (high school) does improve high school performance over \fedavg\ (uniform), it results in drastically worse accuracy on the census domain. This is somewhat expected as the number of census clients far outsizes the number of high school clients. \afa\ outperforms \fedavg\ (uniform) on the high school domain and also significantly decreases the gap in accuracy between high school and census.

\begin{table}[h]
\centering
\caption{Statistics per domain in the EMNIST-62 dataset.}
\small{
\begin{tabular}{ccc}
                        & train & test  \\ \hline
high school clients       & 500   & 500   \\
census clients            & 2900  & 2900  \\
high school examples    & 68.8K & 8.7K  \\
census examples         & 597K  & 74.4K
\end{tabular}
\label{table:emnist-data}
}
\end{table}
\begin{table}[h]
\caption{Accuracy for EMNIST-62 test dataset with the standard deviation for three trials in parentheses.}
\centering
\small{
\begin{tabular}{cccc}
algorithm               & \textbf{high school}  & census                & difference    \\ \hline
\fedavg\ (uniform)      & $82.6 (1.0)$          & $\mathbf{86.3 (0.5)}$ & $3.7$         \\
\fedavg\ (high school)  & $\mathbf{88.2 (0.3)}$ & $75.1 (0.2)$          & $13.1$        \\
\afa                    & $\mathbf{85.7 (0.9)}$ & $84.9 (1.1)$          & $\mathbf{0.8}$

\label{table:emnist-results}
\end{tabular}
}
\end{table}

\end{document}